\documentclass[11pt]{videococo}
\usepackage{xspace}
\usepackage{microtype}

\newcommand{\method}{PrimiTok\xspace}
\newcommand{\myparagraph}[1]{\vspace{0.5em}\noindent\textbf{#1}}

\definecolor{primitokrow}{RGB}{231,241,250}
\makeatletter
\newcommand{\primitokrowbackground}{%
  \leavevmode\rlap{%
    \hspace{-\tabcolsep}%
    \color{primitokrow}%
    \rule[-\dp\@arstrutbox]{\dimexpr\linewidth+2\tabcolsep\relax}%
      {\dimexpr\ht\@arstrutbox+\dp\@arstrutbox\relax}%
  }%
}
\makeatother
\definecolor{top1orange}{RGB}{244,165,66}
\definecolor{top2orange}{RGB}{247,231,206}

\newfontfamily\arxivkeepbytesansmedium[Path=./]{ByteSans-Medium.ttf}
\newfontfamily\arxivkeepbytesansbold[Path=./]{ByteSans-Bold.ttf}
\newsavebox{\arxivdependencybox}
\AtBeginDocument{%
  \sbox{\arxivdependencybox}{%
    \includegraphics{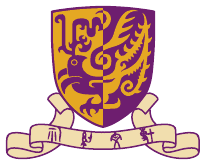}%
    
    \includegraphics{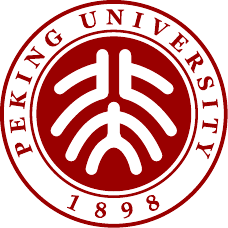}%
  }%
}

\title{\gradtext{PrimiTok}: Learning to Diversify and Allocate Primitives for Visual Tokenization}

\author[1]{Yizhu Chen}
\author[2,\ddagger]{Chen Ju}
\author[3]{Donghao Zhou}
\author[3]{Ziyu Guo}
\author[2]{Shuai Xiao}
\author[2]{Jinsong Lan}
\author[2]{Xiaoyong Zhu}
\author[2]{Ying Chen}
\author[3,\dagger]{Haodong Li}

\contribution[\ddagger]{Project leader}
\contribution[\dagger]{Corresponding author.}

\affiliation[1]{Peking University}
\affiliation[2]{Alibaba Group}
\affiliation[3]{The Chinese University of Hong Kong}

\abstract{
Unifying visual understanding and generation requires representations that preserve both semantic information and fine-grained image details.
Continuous tokenizers trained primarily for semantic alignment may omit fine-grained reconstruction details, whereas discrete tokenizers face information loss from codebook quantization.
Conventional multi-codebook designs assign every token the same number of codewords regardless of local content.
We instead view codewords as reusable visual primitives whose allocation and composition adapt to each token.
We introduce PrimiTok, a compositional visual tokenizer that represents each visual token through adaptive allocation and weighted composition of learned codebook primitives. PrimiTok integrates two complementary components: Diverse Quantitative Primitives (DQP), which encourages diversity among sub-codebook centroids to reduce redundancy, and a Dynamic Primitive Allocator (DPA), which selects primitives according to token complexity. Embedding quantization error supervises how many primitives are selected from each sub-codebook, and DPA aggregates them with input-dependent weights.
Joint reconstruction and image--text contrastive learning encourage the resulting representations to retain visual fidelity and semantic alignment. Experiments demonstrate improved reconstruction and semantic alignment over SOTA baselines, with further evaluations on multimodal understanding and visual generation.
}

\date{November 2026}

\begin{document}
\maketitle

% ── Sections ─────────────────────────────────────

\section{Introduction}

% gemini精简
% In recent years, the rapid development of large-scale models (LMs) has swept through the artificial intelligence community, reshaping productivity everywhere and demonstrating transformative impacts across a broad spectrum of downstream applications. Generally speaking, the capabilities of LMs can be roughly divided into two fundamental groups: understanding and generation. This dichotomy has given rise to several architectural branches, {\em e.g.}, Large Language Models (LLMs) that excel in text understanding and generation, diffusion models (DFMs) that pioneer high-fidelity image generation, and multimodal large language models (MLLMs) that integrate image-text understanding.

% The rapid advancement of large-scale models (LMs) has significantly advanced artificial intelligence, transforming various applications. LMs broadly exhibit two crucial capabilities: understanding and generation. This has led to distinct architectural branches, including Large Language Models (LLMs) for the text domain, diffusion models (DFMs) for high-fidelity image generation, and multimodal large language models (MLLMs) for integrated image-text understanding.

Unified multimodal models (UMMs) aim to integrate multimodal understanding and generation within a shared modeling framework~\citep{wu2024vila,zhou2025transfusion}.
Visual tokenization, which is a key component of UMMs, provides the interface between language and images, transforming visual content into representations for joint modeling with text.
A tokenizer serving both tasks should preserve semantic information for cross-modal alignment and reasoning, as well as fine-grained details for faithful image reconstruction~\citep{ma2025unitok,zhao2025qlip,yue2025uniflow}.

% In NLP, LLMs have largely unified understanding and generation, while multimodal systems remain divided between MLLMs and diffusion foundation models. Vision tokenization is therefore key to multimodal unification, allowing language models to process and generate both visual and textual tokens, as shown in Figure~\ref{fig: introduction}. Existing methods mainly adopt continuous tokenization (CT)~\citep{deng2025bagel,dong2023dreamllm,ge2023planting,li2024mini} or discrete tokenization (DT)~\citep{janus_Wu_2025_CVPR,wu2024vila,ma2025unitok}. CT leverages pretrained MLLMs for understanding and external diffusion modules for generation, achieving strong performance at the cost of complex pipelines and substantial engineering overhead. DT instead quantizes images into discrete tokens that can be modeled jointly with text by a single MLLM, but information loss from codebook discretization often limits performance despite advances in vector quantization~\citep{VQVAE_NIPS2017_7a98af17,VQGAN_2021_CVPR,yu2021vector,RQVAE_Lee_2022_CVPR}.
Existing methods mainly adopt continuous tokenization (CT)~\citep{deng2025bagel,dong2023dreamllm,li2024mini} or discrete tokenization (DT)~\citep{ge2023planting,wu2024vila,ma2025unitok}.
A common CT-based design combines pretrained visual encoders with LLMs for understanding and uses an external diffusion model conditioned on LLM outputs for image generation~\citep{dong2023dreamllm,sun2023emu}.
Continuous features optimized for semantic alignment may omit fine-grained information needed for faithful reconstruction, particularly with a frozen visual encoder~\citep{ma2025unitok,wu2024vila,yue2025uniflow}. DT instead quantizes images into discrete tokens that can be modeled jointly with text under a shared next-token prediction objective.  Preserving task-relevant information during codebook discretization is essential for both understanding and generation~\citep{VQVAE_NIPS2017_7a98af17,ma2025unitok}, motivating advances in vector quantization~\citep{VQGAN_2021_CVPR,yu2021vector,RQVAE_Lee_2022_CVPR}.

% Existing methods mainly adopt continuous tokenization (CT)~\citep{deng2025bagel,dong2023dreamllm,ge2023planting,li2024mini} or discrete tokenization (DT)~\citep{janus_Wu_2025_CVPR,wu2024vila,ma2025unitok}.
% \textcolor{blue}{Continuous tokenizers retain real-valued features, whereas discrete tokenizers map visual features to a finite set of codes, introducing quantization error~\citep{VQVAE_NIPS2017_7a98af17,VQGAN_2021_CVPR}.}
% \textcolor{blue}{These representation choices should be distinguished from downstream model architectures: for example, Transfusion jointly models text and continuous image representations within a single Transformer using language-modeling and diffusion objectives~\citep{zhou2025transfusion}.}
% \textcolor{blue}{Figure~\ref{fig: introduction} illustrates representative tokenization pipelines.}
% \textcolor{blue}{Our focus is the flexibility of the visual representation, particularly how codebook primitives are combined for each token.}

\begin{figure*}[t]
\centering
\vspace{-0.5cm}
\includegraphics[width=1.0\linewidth]{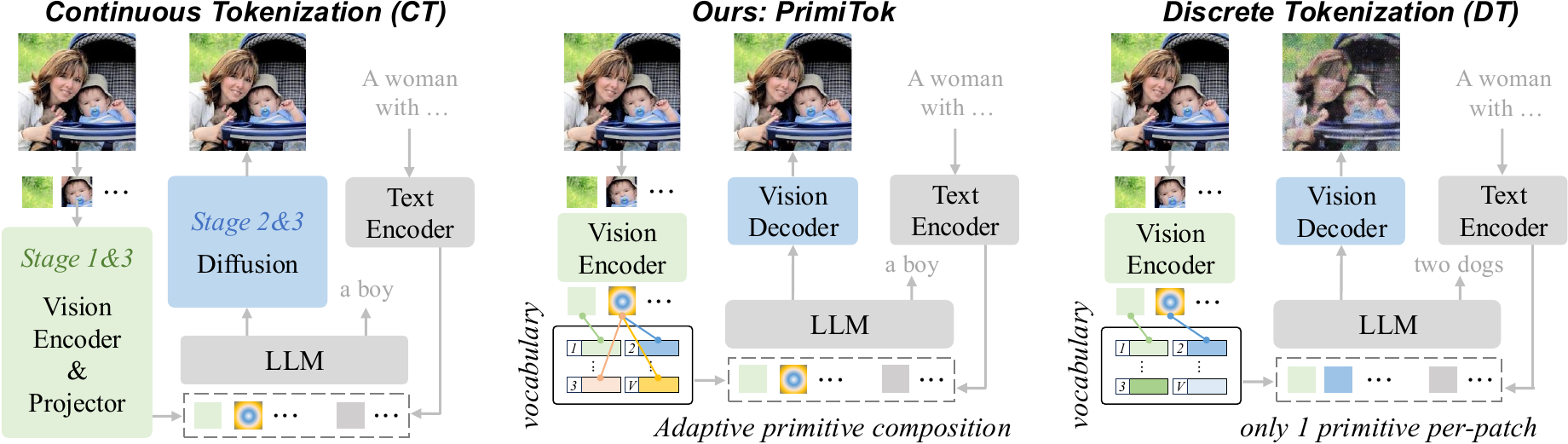}
\vspace{-0.6cm}
% \caption{\textbf{Tokenization Comparisons.} \textbf{\textit{Continuous Tokenizers (CT)}}: strong performance, but one complex, multi-stage workflow. \textbf{\textit{Discrete Tokenizers (DT)}}: a concise-unified workflow, but poor results. \textbf{\textit{\method}} as continuous-discrete dualistic quantization, gets remarkable results in elegant unity of understanding \& generation.}
\caption{\textbf{Tokenization Comparisons.}
CT trained primarily for semantic alignment may omit fine-grained reconstruction details and often rely on auxiliary models for high-quality image synthesis.
DT can lose information through codebook quantization.
\textit{\method} adapts primitive allocation and composition, using reconstruction and image--text contrastive learning to preserve fidelity and semantics.}
\label{fig: introduction}
\vspace{-0.5cm}
\end{figure*} 

% To address the inherent tension: maintaining a concise-scalable framework while pursuing strong performance, we examine a fundamental question: 
% \underline{must visual tokenization be limited to the binary} \underline{choice between discrete or continuous?} \textit{Drawing an analogy from the wave–particle duality of light in physics—where light exhibits both continuous (wave) and discrete (particle) characteristics—we propose a continuous–discrete dualistic visual tokenizer (\method) that bridges understanding and generation by allowing each visual token to be represented at an adaptive level of discretization.} 
% This motivates us to revisit not only the codewords that are learned, but also how they are selected and combined to represent each visual token.
% We introduce \textbf{\method}, a compositional visual tokenizer that treats codebook entries as reusable primitives. 
% Within each sub-codebook, \method selects a content-dependent number of primitives and aggregates them with input-dependent, real-valued weights, as illustrated in Figure~\ref{fig: introduction}.

This motivates us to view codebook entries as reusable visual primitives and revisit not only which primitives are learned, but also how they are selected and combined to represent each visual token.
We introduce \textbf{\method}, a compositional visual tokenizer that learns content-adaptive primitive allocation.
\method expands the expressive capacity of a finite primitive vocabulary through composition, while adapting the number of selected primitives within each sub-codebook to each token's representational needs (Figure~\ref{fig: introduction}).

We base this design on two principles. First, a weighted combination of codebook primitives can express variations that are not captured by individual codewords, providing a more flexible approximation of continuous visual features. Second, the number of primitives should adapt to the representational needs of each token rather than remain fixed. Tokens that are well approximated by a few primitives can use compact compositions, while others can draw on richer combinations. \method thus retains a finite primitive vocabulary while adapting both the number of selected primitives and their composition weights to the input content.

In response to these principles, we develop \textit{Diverse Quantitative Primitives (DQP)} and a \textit{Dynamic Primitive Allocator (DPA)} within an encoder-quantizer-decoder framework.
Specifically, DQP regularizes pairwise similarity between sub-codebook centroids to encourage diversity and reduce redundancy in the primitive vocabulary. DPA takes image features as input and uses embedding quantization error as a training signal for token complexity; it predicts an allocation ratio, retains relevant primitives, and aggregates them with input-dependent weights. The two modules address complementary aspects of representation: DQP shapes the available building blocks, while DPA determines their content-dependent use. Together, they provide diverse candidates and adapt their allocation to each token's approximation difficulty.
An image decoder reconstructs the input from the composed embeddings, while a text encoder provides image-text alignment supervision. Following the joint reconstruction and semantic-alignment paradigm, we optimize both objectives to encourage visual fidelity and semantic retention.

% We conduct extensive experiments on image classification, image–text retrieval, image reconstruction, and image generation, demonstrating that \method achieves comparable or superior performance relative to discrete and continuous tokenizers. Specifically, on ImageNet, \method achieves an impressive 0.31 reconstruction FID and 70.5\% zero-shot Top-1 accuracy at 256$\times$256 resolution, showing that a shared compositional representation can support both high-fidelity reconstruction and semantic discrimination.

We evaluate \method on zero-shot image classification, image--text retrieval, image reconstruction, multimodal understanding, and visual generation. Compared with matched UniTok baselines, \method improves both reconstruction fidelity and semantic alignment.
On ImageNet-1K at $256\times256$ resolution, its shared compositional representation supports both high-fidelity reconstruction and semantic discrimination, yielding an rFID of 0.31 and a zero-shot Top-1 accuracy of 70.5\%.
Under the LLaVA-1.5 training protocol, \method also achieves the highest average score on understanding among the state-of-the-art tokenizers.

\section{Related Works}

% \myparagraph{Discrete Image Tokenizers} map visual signals to token sequences. VQVAE \citep{VQVAE_NIPS2017_7a98af17} introduced codebook quantization, and VQGAN \citep{VQGAN_2021_CVPR} added GAN loss for better reconstruction. Recent models like LlamaGen~\citep{sun2024llamagen}, VAR~\citep{VAR_NEURIPS2024_9a24e284}, and UniTok~\citep{ma2025unitok} show success in autoregressive (AR) generation. Beyond fixed-length designs, recent works explore variable-length tokenization. \citet{jin2024unified} merges patches to shorten discrete sequences. ALIT~\citep{duggal2024adaptive} adjusts representation length through recurrent token allocation, while DOVE~\citep{mao2025DOVE} produces variable-length continuous tokens. However, these methods still lack the fine-grained tokens needed for high-fidelity generation and robust understanding.
% Due to the substantial information loss induced by conventional quantization, the performance upper bound of AR methods may be limited, as evidenced in~\citet{janus_Wu_2025_CVPR, showo_xie2024show}. In our work, we address this limitation by using adaptive granularity quantization to dynamically generate finer-grained tokens, without merging features before quantization, which substantially improves reconstruction quality while maintaining compatibility with AR generation.

\myparagraph{Discrete Image Tokenizers} map visual signals to sequences of discrete codebook assignments. VQ-VAE~\citep{VQVAE_NIPS2017_7a98af17} learns discrete latent representations by mapping encoder outputs to their nearest codebook entries. VQGAN~\citep{VQGAN_2021_CVPR} builds on this approach with perceptual and adversarial losses to improve the perceptual quality of reconstructed images. Building on discrete visual tokens, LlamaGen~\citep{sun2024llamagen}, VAR~\citep{VAR_NEURIPS2024_9a24e284}, and UniTok~\citep{ma2025unitok} support autoregressive (AR) image generation. Beyond fixed-length designs, \citet{jin2024unified} merges patches to shorten discrete sequences, while ALIT~\citep{duggal2024adaptive} adjusts representation length through recurrent token allocation. These adaptive-length methods vary the output sequence length. In contrast, \method reduces quantization loss through content-adaptive primitive allocation and weighted composition within each visual token.

% \myparagraph{Continuous Image Tokenizers} map image patches to continuous latent tokens, usually with less information loss than discrete codes. They typically use an encoder and an MLP or a linear projector to compress images into a latent space. Most models rely on diffusion heads to model the token distribution. MAR~\citep{MAR_NEURIPS2024_66e22646} uses diffusion loss for autoregressive generation. Fluid~\citep{fan2024fluid} scales this paradigm to text-to-image tasks. MetaMorph~\citep{tong2024metamorph} predicts continuous SigLIP~\citep{SigLip_2023_ICCV} embeddings. Finally, FAR~\citep{FAR_yu2025frequency} adds causality and spatial locality to improve AR training efficiency.

\myparagraph{Continuous Image Tokenizers} encode images as real-valued visual features rather than discrete codeword assignments. These features can be adapted to downstream models through an MLP or linear projector. DOVE~\citep{mao2025DOVE} produces variable-length sequences of continuous visual tokens. At the generation stage, MAR~\citep{MAR_NEURIPS2024_66e22646} models continuous tokens autoregressively with a diffusion loss, and Fluid~\citep{fan2024fluid} extends this approach to text-to-image generation. MetaMorph~\citep{tong2024metamorph} autoregressively predicts continuous SigLIP~\citep{SigLip_2023_ICCV} embeddings, while FAR~\citep{FAR_yu2025frequency} models continuous tokens in a frequency-progressive order to preserve causality and spatial locality.

\myparagraph{Unified Multimodal Models} integrate visual understanding and generation within a shared framework. Representative systems use continuous visual representations with diffusion-based image generation~\citep{dong2023dreamllm,li2024mini,sun2023emu}. Other systems model discrete visual and text tokens autoregressively under a shared next-token prediction objective~\citep{ge2023planting,wu2024vila,wu2024liquid,wang2024emu3,ma2025unitok,zhao2025qlip}. Diffusion-based decoding can complement autoregressive visual-token prediction for either discrete or continuous representations~\citep{ge2023planting,tong2024metamorph}. We compare \method with representative visual tokenizers used in these systems.

% Codebook Filter module codebook diversity
% This work proposes a novel method for adaptive quantization. 

\section{Method: \method}
\begin{figure*}[t]
\vspace{-0.5cm}
\centering
\includegraphics[width=1.0\linewidth]{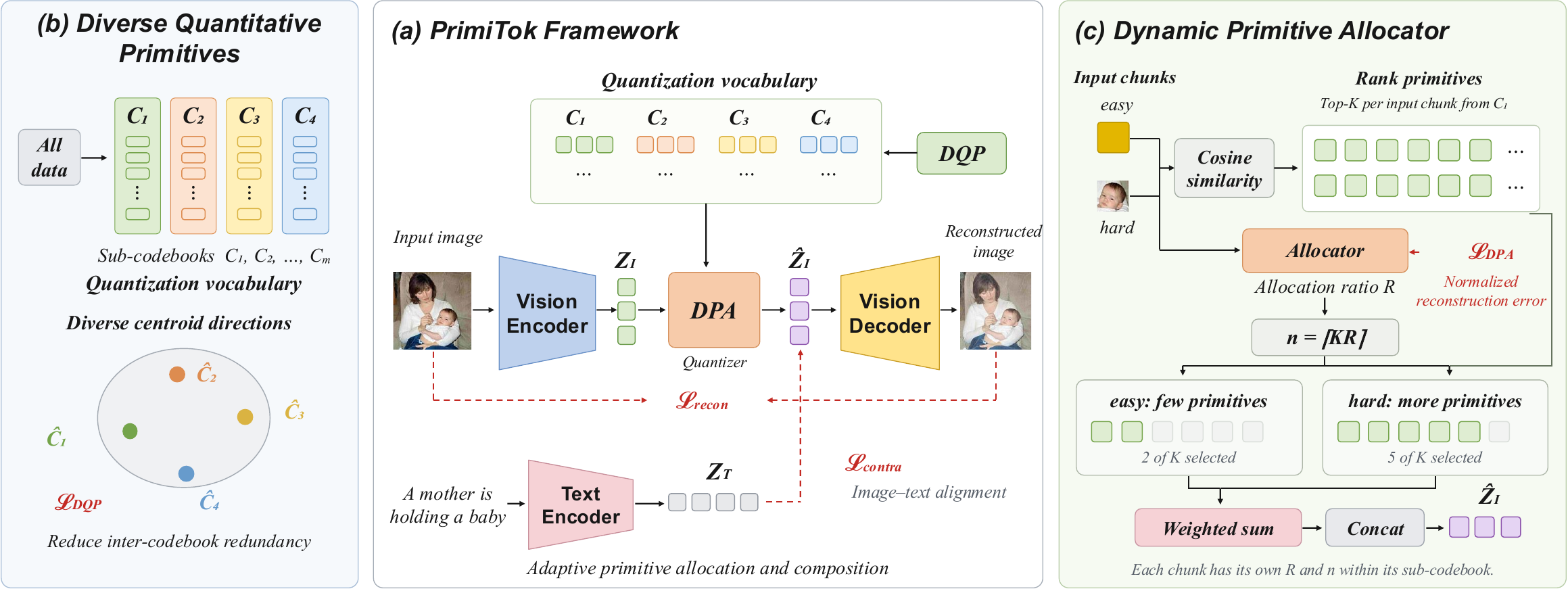}

% \vspace{-0.2cm}
% \caption{\textbf{Framework Overview.} \method consists of an image encoder, a text encoder, an image decoder, and two core components: Diverse Quantitative Primitives (DQP) and Dynamic Primitive Allocator (DPA). DQP encourages centroid diversity among sub-codebooks, while DPA assesses token complexity to allocate relevant primitives.}
\caption{\textbf{Framework Overview.} \method consists of an image encoder, a text encoder, an image decoder, and two core components, Diverse Quantitative Primitives (DQP) and a Dynamic Primitive Allocator (DPA).
(a) We train \method through joint image reconstruction and image--text contrastive learning.
(b) DQP encourages centroid diversity across sub-codebooks to reduce redundancy.
(c) DPA adapts primitive counts and composition weights to each token chunk. The quantized chunks are concatenated into the final image embedding.}
\label{fig: framework}
\vspace{-0.5cm}
\end{figure*}

% For the unification of understanding and generation, the key is to maintain concise-scalable framework while pursuing strong performance. Hence, we propose a Continuous-Discrete Dualistic Visual Tokenizer (\method). In Sec.~\ref{sec:formulation}, we first outline formulation and framework; in Sec.~\ref{sec:DQP}, we pursue orthogonal image primitives; in Sec.~\ref{sec:DPA}, we adaptively allocate the optimal set of primitives by information complexity; in Sec.~\ref{sec: training}, we summarize training stages of \method and all optimization objectives. An overview of the framework is presented in Figure~\ref{fig: framework}.
% To unify visual understanding and generation, we aim to maintain a concise and scalable framework while achieving strong performance. 
To unify visual understanding and generation, we aim to preserve both semantic information and fine-grained visual details in a shared representation. To this end, we propose PrimiTok. In Sec.~\ref{sec:formulation}, we outline the formulation and framework; in Sec.~\ref{sec:DQP}, we encourage diversity among sub-codebook centroids; in Sec.~\ref{sec:DPA}, we adaptively allocate relevant primitives according to token complexity; in Sec.~\ref{sec: training}, we summarize training stages of PrimiTok and optimization objectives. Figure~\ref{fig: framework} illustrates the framework.

\subsection{Framework Overview} 
\label{sec:formulation}

% \textcolor{red}{What problems does the image tokenizer need to solve, and what are the input and output?}
% 这段主要讲，对于VQ-based image tokenizers，输入输出是什么，训练和测试阶段的目标。
% 讲我们的方法的框架 以及完成classification retrieval 等任务的输入输出和用到的模块

We first present the training pipeline of \method, which learns from image--text pairs. Following an encoder--quantizer--decoder architecture~\citep{VQVAE_NIPS2017_7a98af17}, \method comprises an image encoder $\mathcal{E}_\text{I}$, a quantizer $\mathcal{Q}$, and an image decoder $\mathcal{D}$. The quantizer integrates the \textbf{Diverse Quantitative Primitives} $\mathcal{Q}_\text{DQP}$ and the \textbf{Dynamic Primitive Allocator} $\mathcal{Q}_\text{DPA}$, while a text encoder $\mathcal{E}_\text{T}$ provides text embeddings for image--text contrastive learning.

% During \textbf{tokenization}, for one input image $\mathbf{I} \in \mathbb{R}^{H' \times W' \times C}$, we first partition it into patches, and then pass patches through $\mathcal{E}_\text{I}$ into latent embeddings $\mathbf{Z}_\text{I} = \mathcal{E}_\text{I}(\mathbf{I}) \in \mathbb{R}^{hw \times D}$, where $D$ is the embedding dimension and $H/W$ denotes the downsampling of the spatial dimensions $H'/W'$. For one input text $\mathbf{T} \in \mathbb{R}^{N}$ of $N$ words, we obtain its latent embeddings by $\mathbf{Z}_\text{T} = \mathcal{E}_\text{T}(\mathbf{T}) \in \mathbb{R}^{N \times D}$.

% During \textbf{tokenization}, the vision encoder $\mathcal{E}_\text{I}$ projects an image $\mathbf{I} \in \mathbb{R}^{H \times W \times C}$ into latent embeddings $\mathbf{Z}_\text{I} \in \mathbb{R}^{hw \times D}$. These $hw$ visual tokens are $D$-dimensional vectors, each representing a local image patch. Similarly, text $\mathbf{T}$ is encoded into $N$ language tokens $\mathbf{Z}_\text{T} \in \mathbb{R}^{N \times D}$.

During \textbf{tokenization}, the vision encoder $\mathcal{E}_\text{I}$ projects an image $\mathbf{I} \in \mathbb{R}^{H \times W \times C}$ into latent embeddings $\mathbf{Z}_\text{I} \in \mathbb{R}^{hw \times D}$. Here, $H$, $W$, and $C$ denote image height, width, and channel count, while $h$ and $w$ denote the latent grid dimensions. These $hw$ visual tokens are $D$-dimensional vectors, each representing a local image patch. Similarly, text $\mathbf{T}$ is encoded into $N$ language tokens $\mathbf{Z}_\text{T} \in \mathbb{R}^{N \times D}$.

During \textbf{quantization}, $\mathcal{Q}_\text{DQP}$ maintains $M$ sub-codebooks $\mathbf{C}$ composed of diverse primitives. For image embeddings $\mathbf{Z}_\text{I}$, $\mathcal{Q}_\text{DPA}$ assesses the information complexity of each token and adaptively allocates primitives accordingly. This yields the quantized image embeddings $\hat{\mathbf{Z}}_\text{I}$, which we align with the text embeddings $\mathbf{Z}_\text{T}$.

During \textbf{de-tokenization}, $\mathcal{D}$ reconstructs image $\hat{\mathbf{I}}$ from $\hat{\mathbf{Z}}_\text{I}$, and the overall formulation is: 
% \begin{align}
% 	 & \resizebox{0.88\linewidth}{!}{$\hat{\mathbf{I}} = \mathcal{D}(\hat{\mathbf{Z}}_{\text{I}}) = \mathcal{D}(\mathcal{Q}_{\text{DPA}}(\mathbf{Z}_{\text{I}})) = \mathcal{D}(\mathcal{Q}_{\text{DPA}}(\mathcal{E}_\text{I}(\mathbf{I})))$,}\\
%    & \{\mathbf{C}_j\}_{j=1}^{M} = \mathcal{Q}_{\text{DQP}}(\mathbf{I}).
% \end{align}

\begin{gather}
\hat{\mathbf{I}} = \mathcal{D}(\hat{\mathbf{Z}}_{\text{I}}) = \mathcal{D}(\mathcal{Q}_{\text{DPA}}(\mathbf{Z}_{\text{I}})) = \mathcal{D}(\mathcal{Q}_{\text{DPA}}(\mathcal{E}_\text{I}(\mathbf{I}))),\\
\{\mathbf{C}_j\}_{j=1}^{M} = \mathcal{Q}_{\text{DQP}}(\mathbf{I}).
\end{gather}

\subsection{Diverse Quantitative Primitives}
\label{sec:DQP}
For primitive quantization, one typical solution is codebook construction~\citep{VQVAE_NIPS2017_7a98af17, VQGAN_2021_CVPR}, {\em i.e.}, modeling data distribution to learn primitive embeddings that serve as the basic units for decomposing continuous visual tokens. The challenge lies in minimizing information loss. Here, we encourage a more diverse distribution of primitive embeddings to improve coverage of the continuous information space.

Inspired by multi-head attention~\citep{NIPS2017_attention}, our \textit{multi-head codebook structure} promotes primitive diversity. We define $M$ sub-codebooks $\mathbf{C}_j = \{ \mathbf{c}_{i,j} \}_{i=1}^{v} \in \mathbb{R}^{v \times d}$ each with a local vocabulary size $v$. Here, $\mathbf{c}_{i,j}$ denotes the $i$-th primitive vector in sub-codebook $j$, and $d$ is the chunk dimension. These $M$ heads jointly form a higher-dimensional representation $D = M \times d$, expanding the global vocabulary through various primitive combinations. During quantization, we split each visual token along the feature dimension into $M$ chunks; $\mathbf{Z}_{\text{I},j} \in \mathbb{R}^{hw \times d}$ collects the $j$-th chunk of every token, for $j=1,\ldots,M$. We quantize each chunk independently using its corresponding sub-codebook. This approach enables a highly flexible and diverse representation within the expanded latent space.

\begin{table*}[t]
\centering
\small 
\vspace{-0.5cm}
  \caption{\textbf{Image Reconstruction and Zero-Shot Image Classification on ImageNet-1K 50K Validation Set.} $^\dag$ indicates CLIP pre-trained initialization. $^\#$ indicates VQGAN pre-trained initialization. $^*$ denotes our reproduction.}
  \label{tab: imagenet}
  \vspace{-0.2cm}
  
\footnotesize
\setlength{\tabcolsep}{3pt}
\renewcommand{\arraystretch}{1.15}
\begin{tabular*}{\linewidth}{@{\extracolsep{\fill}}l c c c c c c c c@{}}
    \toprule
\textbf{Method} & \textbf{Data} &
\textbf{\#Tokens} & 
 \textbf{Res.} & 
   \textbf{Codebook Size} &
\textbf{Accuracy} & 
\textbf{rFID$\downarrow$} & 
   \textbf{PSNR$\uparrow$} & 
    \textbf{LPIPS-VGG$\downarrow$} \\
   \midrule
  \multicolumn{9}{c}{\textit{Continuous Tokenizer (CT)}} \\
   \midrule
     EVA02-B/16 & Merged-2B & -- & 224 & -- &  74.7    &  --     &   -- &   --  \\
    CLIP-L/14 & WIT400M & -- & 224 & -- & 75.5 &  --    &  -- &   -- \\
    MetaCLIP-L/14 & MetaCLIP400M & -- & 224 & -- & 76.2 &  --  &   -- &   -- \\
         DOVE$^{\#}$    &ImageNet-1K  & 256 & 256 & 8192 &  -- &  7.73 & -- & --\\
      \midrule
  \multicolumn{9}{c}{\textit{Discrete Tokenizer (DT)}} \\
   \midrule
    VQGAN & ImageNet-1K & 256  & 256 & 16384 &  --     &   4.98 &   20.00  &0.2843  \\
    SD-VQGAN & OpenImages & 256  & 256 & 16384 &  --     &   5.15 &  20.83   & -- \\
    LlamaGen & ImageNet-1K & 256 & 256 & 16384 &  --     &   2.19 &   20.79  &  -- \\
    $\text{TokenFlow}^{\dag}$  & LAION+COYO700M & 256 & 256 &  32768  & -- &1.37 & 21.41   & --  \\
    $\text{VILA-U}^{\dag}$  & COYO700M & 256  & 256 & 16384 & 73.3 &   1.80  &   -- &   --\\
    TiTok-S & MaskGIT-VQGAN & 128 & 256   & 4096 & --   &   1.71    &  --      &   --    \\
    QLIP$^{\dag}$      & DC-1B & -- & 256 & -- & 74.3 &  3.21 & 23.16 & --\\
       \midrule
   Encoder-Only  & DC-800M & --  & 256 & -- & 66.8 &   --  &   --    &   --   \\
    $\text{UniTok}^*$  & DC-800M & 256 & 256 & 32768 &  65.8 &  0.41 & 23.57   & 0.1440 \\
    \rowcolor{primitokrow} \primitokrowbackground \textbf{\method} & DC-800M  & 256 & 256 & 32768 & 66.5  &  0.37  &   23.93  &  0.1410 \\
     % $\text{UniTok}^\dag$  & DC-800M & 256 & 256 & 32768 &  69.1 &  0.45 & 24.46    & 0.1344\\
    % \makecell[c]{60\%\\DC-1B}
       \midrule
   Encoder-Only  & DC-1B-R & --  & 256 & -- & 70.1 &   --  &   --    &   --   \\
      UniTok      & DC-1B & 256 & 256 & 32768 & 70.8 &  0.41 & 23.80   & 0.1458  \\
        % $\text{UniTok}^*$      & DC-1B-R & 256 & 256 & 32768 &  &   &   &   \\
       UniTok$^{*}$     & DC-1B-R & 256 & 256 & 32768 & 68.6 & 0.41  &  23.90   & 0.1366  \\
    \rowcolor{primitokrow} \primitokrowbackground \textbf{\method} & DC-1B-R & 256 & 256 & 32768 &  69.8    &  0.36    &   24.51  &   0.1289  \\ 
       \midrule
           $\text{UniTok}^{\dag}$  & DC-1B & 256  & 256 & 32768 & 78.6 & 0.38   &      -- &   --   \\
      $\text{UniTok}^{\dag*}$  & DC-1B-R & 256  & 256 & 32768 & 69.9 & 0.35   &      24.40 &   0.1288   \\
    \rowcolor{primitokrow} \primitokrowbackground \textbf{\method}$^{\dag}$ & DC-1B-R & 256 & 256 & 32768 &  70.5    &   0.31    &   24.95     &   0.1197    \\
    \bottomrule
  \end{tabular*}

\vspace{-0.3cm}
\end{table*}

% To ensure the $M$ sub-codebooks remain distinct, we introduce a diversity loss during optimization. Specifically, we minimize the pairwise cosine similarity ($\Phi_\text{sim}$) between sub-codebook centroids. For each sub-codebook $j$, the centroid $\hat{\mathbf{C}}_j$ is defined as the average of all primitives: $\hat{\mathbf{C}}_j = \frac{1}{v} \sum_{i=1}^{v} \mathbf{c}_{i,j}$. This discourages primitive homogenization, where different heads learn redundant information. The regularization loss is formulated as:
To mitigate redundancy among the $M$ sub-codebooks, we introduce a diversity loss during optimization. Specifically, we minimize the pairwise cosine similarity $\Phi_\text{sim}$ between sub-codebook centroids. For each sub-codebook $j$, we define the centroid $\hat{\mathbf{C}}_j$ as the average of all primitives: $\hat{\mathbf{C}}_j = \frac{1}{v} \sum_{i=1}^{v} \mathbf{c}_{i,j}$. This discourages primitive homogenization, where different heads learn redundant information. We formulate the regularization loss as:
\begin{align}
\mathcal{L}_{\text{DQP}} = \frac{2}{M(M-1)}\sum_{1 \le j < k \le M} \Phi_\text{sim} \bigl(\hat{\mathbf{C}}_j,\;\hat{\mathbf{C}}_k\bigr),
\end{align} where $M$ is the number of sub-codebooks. Minimizing $\mathcal{L}_{\text{DQP}}$ encourages diverse and potentially complementary centroid directions. This centroid-level regularization aims to reduce redundancy across sub-codebooks.

% \end{align} where $M$ is the number of sub-codebooks. Minimizing $\mathcal{L}_{\text{DQP}}$ forces sub-codebooks to occupy distinct regions in the feature space. This improves the coverage of the resulting discrete primitive set and reduces information loss.

%  有了这些primitives 如何分配。如何度量信息复杂度，先验

% 然后怎么决定分那个集合。

\subsection{Dynamic Primitive Allocator}
\label{sec:DPA}
With diverse primitives established, we aim to reduce quantization loss. This involves determining both the number of primitives to use and which ones to select for each token. Conventional discrete tokenization selects one primitive per sub-codebook for each token, potentially losing task-relevant information. In contrast, DPA adaptively allocates primitives according to token complexity. This approach reduces quantization error through content-adaptive primitive composition.

\myparagraph{Content-Adaptive Primitive Allocation.} To select primitives from the $j$-th sub-codebook $\mathbf{C}_j$, we use the relevance function $\Phi_{\text{rel}}$ to score input chunks $\mathbf{Z}_{\text{I},j}$ against its primitives via cosine similarity. The \textit{allocator} $\mathcal{A}$ uses an independent branch $\mathcal{A}_j$ to predict a vector of allocation ratios $\mathbf{R}_j = \mathcal{A}_j(\mathbf{Z}_{\text{I},j}) \in (0,1]^{hw}$. For token $p=1,\ldots,hw$, its allocation ratio $R_{p,j}$ determines the allocated primitive count $n_{p,j}=\lceil K R_{p,j}\rceil$. The primitive selection operator $\Phi_{\text{top}}$ retains this many highest-scoring primitives from the Top-$K$ candidate pool, where $K$ is the candidate budget. The quantized chunks and final embedding are:
\begin{align}
    \hat{\mathbf{Z}}_{\text{I},j} &= \Phi_{\text{sum}} \Bigl( \Phi_{\text{top}} \bigl( \Phi_{\text{rel}}(\mathbf{Z}_{\text{I},j}, \mathbf{C}_j), \mathbf{R}_j, K \bigr) \Bigr), \label{eq:quant_chunk} \\
    \hat{\mathbf{Z}}_{\text{I}} &= \text{Concat}(\hat{\mathbf{Z}}_{\text{I},1}, \hat{\mathbf{Z}}_{\text{I},2}, \dots, \hat{\mathbf{Z}}_{\text{I},M}), \label{eq:concat}
\end{align} where $\Phi_{\text{top}}$ and the weighted aggregation operator $\Phi_{\text{sum}}$ operate independently on each token. We aggregate the selected primitives using softmax weights computed from their negative Euclidean distances to the input chunk, with both the chunk and primitives L2-normalized. $\text{Concat}$ denotes concatenation along the feature dimension. Each allocator branch $\mathcal{A}_j$ comprises two 1D convolutional layers followed by a sigmoid activation.

\myparagraph{Measurement of Information Complexity.} We adopt the intuitive prior that higher reconstruction error suggests greater token complexity, as regions with higher information complexity are more prone to information loss during tokenization-quantization-detokenization. 

We define the chunk \textit{reconstruction error} in the embedding space~\citep{VQVAE_NIPS2017_7a98af17} as $e_{p,j} = \frac{1}{d}\lVert \hat{\mathbf{z}}_{p,j} - \bar{\mathbf{z}}_{p,j} \rVert_2^{2}$, where $\hat{\mathbf{z}}_{p,j}$ denotes the $p$-th quantized chunk in $\hat{\mathbf{Z}}_{\mathrm{I},j}$, and $\bar{\mathbf{z}}_{p,j}$ denotes the L2-normalized input chunk from $\mathbf{Z}_{\mathrm{I},j}$. These errors form the reconstruction-error vector $\mathbf{e}_j = (e_{1,j},\ldots,e_{hw,j})^{\top}$. The target-ratio mapping $\Phi_{\text{norm}}$ robustly normalizes these errors to $[1/K,1]$, independently for each image and sub-codebook. We supervise the allocator with the loss
\begin{equation}
\mathcal{L}_{\text{DPA}} = \frac{1}{M}\sum_{j=1}^{M}\operatorname{MSE}\bigl(\mathbf{R}_j,\Phi_{\text{norm}}(\mathbf{e}_j)\bigr),
\label{eq:ratio-target}
\end{equation}
where MSE averages over tokens and no gradient flows through the targets $\Phi_{\text{norm}}(\mathbf{e}_j)$. Thus, simple tokens use one or a few discrete primitives, while complex tokens combine more—up to $K$ per sub-codebook—with input-dependent weights to better approximate continuous visual features.

\begin{figure*}[t]
\vspace{-0.5cm}
\begin{center}
\includegraphics[width=1.0\linewidth]{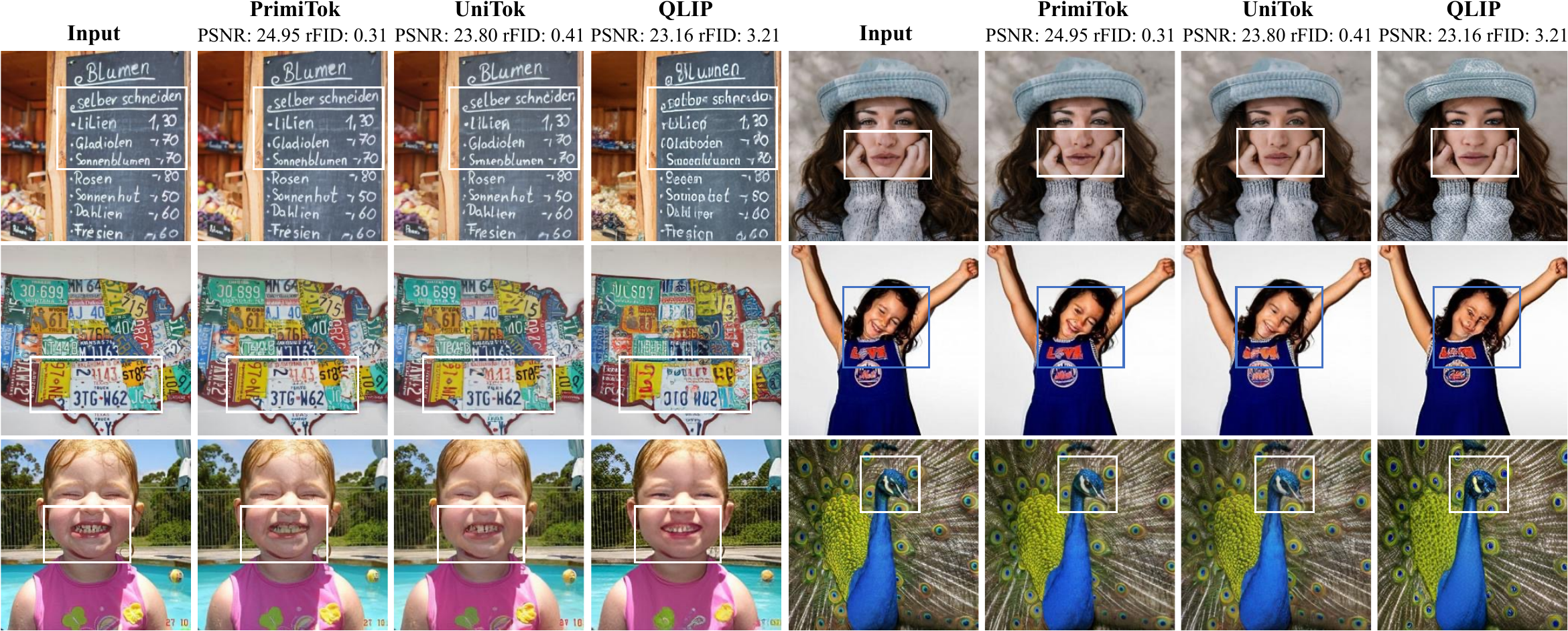}
\end{center}
% \vspace{-0.5cm}
\caption{\textbf{Comparisons of Image Reconstruction between UniTok, QLIP, and \method.} Here, rFID and PSNR are evaluated on ImageNet 50k validation. \method shows superior detail preservation and perceptual fidelity over the competitors, {\em e.g.}, text on the map (Row 2) and blackboard (Row 1) is notably clearer and more faithful to the inputs.}
\vspace{-0.3cm}
\label{fig: reconstruction}
\end{figure*} 

\subsection{Training and Evaluation}
\label{sec: training}

% \textcolor{blue}{To optimize our }
% \begin{equation}
%     \mathcal{L}_{\text{recon}} = \mathcal{L}_{\text{R}} + \lambda_{\text{Q}}\mathcal{L}_{\text{Q}} + \lambda_{\text{P}}\mathcal{L}_{\text{P}} + \lambda_{\text{G}}\mathcal{L}_{\text{G}}
% \end{equation}
% Here, $\mathcal{L}_{\text{R}}$ is the pixel-wise reconstruction loss, $\mathcal{L}_{\text{Q}}$ is the quantization loss between the encoded features and quantized representations, $\mathcal{L}_{\text{P}}$ is the perceptual loss from LPIPS~\citep{LPIPS_2018_CVPR}, and $\mathcal{L}_{\text{G}}$ is the adversarial loss used to improve image quality via a discriminator~\citep{Stylegan_Karras_2019_CVPR}.

\textbf{Training.} Our reconstruction objective $\mathcal{L}_{\text{recon}}$ combines image reconstruction and vector-quantization losses~\citep{VQVAE_NIPS2017_7a98af17} with perceptual and adversarial losses~\citep{VQGAN_2021_CVPR}. The full training objective additionally includes image--text alignment and the DQP and DPA losses. The total loss is thus defined as the $\lambda$-weighted sum of all these components:
\begin{equation}
    \mathcal{L}_{\text{total}} = \mathcal{L}_{\text{recon}} + \lambda_{\text{contra}}\mathcal{L}_{\text{contra}} + \lambda_{\text{DQP}}\mathcal{L}_{\text{DQP}} + \lambda_{\text{DPA}}\mathcal{L}_{\text{DPA}}.
\end{equation} We use InfoNCE~\citep{clip-pmlr-v139-radford21a} for the image--text contrastive loss $\mathcal{L}_{\text{contra}}$, which aligns text $\mathbf{Z}_{\text{T}}$ and quantized image $\hat{\mathbf{Z}}_{\text{I}}$ embeddings. We use a straight-through estimator~\citep{bengio2013estimating} to pass gradients from quantized embeddings to pre-quantization features. We supervise the allocator with $\mathcal{L}_{\text{DPA}}$ and stop gradients through its input features and target ratios.
% $\mathcal{L}_{\text{contra}}$ is the contrastive loss computed via InfoNCE~\citep{clip-pmlr-v139-radford21a} to align text $\mathbf{Z}_{\text{T}}$ and quantized image $\hat{\mathbf{Z}}_{\text{I}}$ embeddings.
% Here, $\mathcal{L}_{\text{R}}$ is the pixel-wise reconstruction loss, $\mathcal{L}_{\text{Q}}$ is the quantization loss between the encoded features and quantized representations, $\mathcal{L}_{\text{P}}$ is the perceptual loss from LPIPS~\citep{LPIPS_2018_CVPR}, and $\mathcal{L}_{\text{G}}$ is the adversarial loss used to improve image quality via a discriminator~\citep{Stylegan_Karras_2019_CVPR}.

\myparagraph{Warm-Up Training.}
DPA relies on reconstruction-error supervision, which may be less reliable early in training (Figure~\ref{fig:ablation_training}). We therefore warm up the tokenizer before enabling content-adaptive allocation. During the first 25\% of training, we disable DQP regularization and DPA-based allocation, quantizing each token chunk to its nearest primitive in the corresponding sub-codebook. We then enable both to promote codebook diversity and content-adaptive primitive allocation.

% \myparagraph{Evaluation.} For understanding tasks, {\em e.g.}, classification and retrieval, we map images and text to embeddings: $\hat{\mathbf{Z}}_{\text{I}} = \mathcal{Q}_{\text{DPA}}(\mathcal{E}_{\text{I}}(\mathbf{I}))$ and $\mathbf{Z}_{\text{T}} = \mathcal{E}_{\text{T}}(\mathbf{T})$. We then rank predictions based on the cosine similarity between $\hat{\mathbf{Z}}_{\text{I}}$ and $\mathbf{Z}_{\text{T}}$, where class names serve as text inputs for classification. For reconstruction, we pass the input image $\mathbf{I}$ through the entire pipeline—encoder, quantizer, and decoder—to obtain the output $\hat{\mathbf{I}} = \mathcal{D}(\hat{\mathbf{Z}}_{\text{I}})$.
\myparagraph{Evaluation.}
For understanding tasks, {\em e.g.}, classification and bidirectional retrieval, we map images and text to embeddings: $\hat{\mathbf{Z}}_{\text{I}} = \mathcal{Q}_{\text{DPA}}(\mathcal{E}_{\text{I}}(\mathbf{I}))$ and $\mathbf{Z}_{\text{T}} = \mathcal{E}_{\text{T}}(\mathbf{T})$.
We then rank predictions based on the cosine similarity between $\hat{\mathbf{Z}}_{\text{I}}$ and $\mathbf{Z}_{\text{T}}$, where class names serve as text inputs for classification.
For reconstruction, we pass the input image $\mathbf{I}$ through the entire pipeline---encoder, quantizer, and decoder---to obtain the output $\hat{\mathbf{I}} = \mathcal{D}(\hat{\mathbf{Z}}_{\text{I}})$.

% \myparagraph{Evaluation.}
% For zero-shot image classification and image--text retrieval, we map images and text to embeddings $\hat{\mathbf{Z}}_{\text{I}} = \mathcal{Q}_{\text{DPA}}(\mathcal{E}_{\text{I}}(\mathbf{I}))$ and $\mathbf{Z}_{\text{T}} = \mathcal{E}_{\text{T}}(\mathbf{T})$.
% We then rank candidate classes or cross-modal matches by the cosine similarity between $\hat{\mathbf{Z}}_{\text{I}}$ and $\mathbf{Z}_{\text{T}}$. We use class names as text inputs for classification and evaluate retrieval in both directions. For reconstruction, we pass the input image $\mathbf{I}$ through through the entire pipeline—encoder, quantizer, and decoder—to obtain the reconstructed image $\hat{\mathbf{I}} = \mathcal{D}(\hat{\mathbf{Z}}_{\text{I}})$.

\section{Experiments}

\subsection{Experimental Setup}

% \subsection{Experimental Setup}

\myparagraph{Datasets and Metrics.}
We evaluate reconstruction on the ImageNet-1K~\citep{imagenet_ILSVRC15} and MSCOCO 2017~\citep{MSCOCO_ECCV} validation sets using rFID~\citep{FID_NIPS2017_8a1d6947}, PSNR, SSIM~\citep{SSIM}, and LPIPS~\citep{LPIPS_2018_CVPR}. We assess semantic alignment via zero-shot ImageNet-1K classification and bidirectional image--text retrieval on the Karpathy test splits~\citep{Karpathy_2015_CVPR} of MSCOCO 2014 and Flickr30K~\citep{Flickr30K}. For retrieval, we report Recall@$K$
\citep{clip-pmlr-v139-radford21a,Lit_2022_CVPR,SigLip_2023_ICCV}. For multimodal understanding, we evaluate on POPE~\citep{li-etal-2023-pope}, GQA~\citep{hudson2019gqa}, TextVQA~\citep{singh2019towards}, MM-Vet~\citep{yu2024mm}, MMBench~\citep{liu2025mmbench}, and MME~\citep{fu2023mme}.

\myparagraph{Tokenizer Configuration.}
\method uses eight sub-codebooks, each with 4,096 primitives of dimension $d=8$, resulting in a latent dimension of $D=64$. The encoder and decoder follow ViTamin-L/16~\citep{vitamin_Chen_2024_CVPR}. Unless otherwise specified, we train for one epoch at $256\times256$ resolution with a batch size of 64 and a learning rate of $10^{-3}$, using either random or CLIP-pretrained initialization. Training data consist of either 0.8B successfully downloaded DataComp-1B~\citep{datacomp} image--text pairs, denoted as DC-800M, or a 1.28B resampled variant, denoted as DC-1B-R, formed by adding 0.48B samples drawn with replacement from DC-800M. Further implementation details are provided in Appendix~1.

% \myparagraph{Tokenizer Setup.} We configure the quantizer with eight sub-codebooks, each containing 4096 code entries and a latent dimension of 8-d, resulting in a global latent dimension of 64-d, following UniTok~\citep{ma2025unitok}. The encoder and decoder are built upon the ViTamin-L/16~\citep{vitamin_Chen_2024_CVPR}. The \textcolor{red}{discriminator} is initialized with pre-trained DINOv2-S~\citep{oquab2023dinov2}. We train \method for one epoch on the public dataset DataComp-1B~\citep{datacomp} consisting of 1.28B image--text pairs\footnote{Due to download issues, we obtained 0.8B pairs, with the remaining 0.48B pairs being resampled.} All images are resized to a 256 $\times$ 256 resolution and a local batch size of 64. The learning rates are set to 1e-3 for the tokenizer and 2e-4 for \textcolor{red}{the discriminator}. For evaluation, we test two initialization settings: CLIP pre-trained weights and default random initialization.

% \footnote{Although larger batch sizes could further improve model performance, we are limited by GPU memory constraints.}

\begin{table*}[t]
\centering
\small
 \vspace{-0.2cm}
\caption{\textbf{Multimodal understanding.}
Baseline results are taken from UniFlow~\citep{yue2025uniflow}. $^\dagger$ denotes the LLaVA-v1.5 training setting. MME-S denotes the sum of perception and cognition scores; MME-P denotes the perception score. Both MME scores are divided by 20 for Avg.}
\label{tab:multimodal_understanding}
\footnotesize
\setlength{\tabcolsep}{3pt}
\renewcommand{\arraystretch}{1.15}
\begin{tabular*}{\linewidth}{@{\extracolsep{\fill}}lllc|cccccccc@{}}
\toprule
\textbf{Method} &
\textbf{Vision Encoder} &
\textbf{LLM} &
\textbf{Res.} &
\textbf{POPE} &
\textbf{GQA} &
\textbf{TQA} &
\textbf{MMV} &
\textbf{MMB} &
\textbf{MME-S} &
\textbf{MME-P} &
\textbf{Avg.} \\
\midrule

\multicolumn{12}{l}{\textit{Understanding-only MLLMs}} \\
LLaVA-1.5
& CLIP-L & Vicuna-7B & 336
& 85.9 & 62.0 & 46.1 & 31.1 & 64.3
& -- & 1510.7 & -- \\

\midrule
\multicolumn{12}{l}{\textit{Unified Tokenizers}} \\

VILA-U$^\dagger$
& SigLIP-SO400M & Vicuna-7B & 256
& 81.6 & -- & -- & -- & --
& -- & 1311.6 & -- \\

UniTok$^\dagger$
& ViTamin-L & Vicuna-7B & 256
& 81.7 & -- & -- & -- & --
& -- & 1448.0 & -- \\

SemHiTok$^\dagger$
& SigLIP-L & Vicuna-7B & 256
& 84.2 & 61.0 & -- & -- & 60.3
& -- & 1400.6 & -- \\

QLIP$^\dagger$
& CLIP-L & Vicuna-7B & 392
& 86.1 & 61.8 & 55.2 & 33.3 & --
& -- & 1498.3 & -- \\

TokenFlow-B$^\dagger$
& CLIP-B & Vicuna-13B & 224
& 84.0 & 59.3 & 49.8 & 22.4 & 55.3
& 1660.4 & 1353.6 & 60.2 \\

TokenFlow-L$^\dagger$
& ViTamin-XL & Vicuna-13B & 256
& 85.0 & 60.3 & 54.1 & 27.7 & 60.3
& 1622.9 & 1365.4 & 62.4 \\

% UniTok
% & ViTamin-L & LLaMA-2-7B & 256
% & 83.2 & 61.1 & 51.6 & 33.9 & --
% & -- & 1448.0 & -- \\

% \midrule
% \multicolumn{12}{l}{\textit{UniFlow-LV: LLaVA-v1.5 setting}} \\

UniFlow-LV$^\dagger$
& DFN-CLIP-L & Vicuna-7B & 224
& 86.5 & 61.3 & 53.4 & 30.2 & 63.8
& 1748.0 & 1446.9 & 65.0 \\

\rowcolor{primitokrow} \primitokrowbackground \textbf{\method}$^\dagger$
& ViTamin-L & Vicuna-7B & 256
& 84.6 & 61.7 & 52.4 & 32.8 & 63.8
& 1787.4 & 1457.1 & 65.4 \\

\bottomrule
\end{tabular*}%

 \vspace{-0.2cm}
\end{table*}

\subsection{Tokenizer Comparisons}
Across three matched training and initialization settings, \method consistently improves ImageNet reconstruction and zero-shot classification over UniTok. Table~\ref{tab: imagenet} further compares \method with representative continuous and discrete tokenizers on the ImageNet-1K 50K validation set~\citep{imagenet_ILSVRC15}. Continuous baselines include EVA02-B/16~\citep{EVA2024105171}, CLIP-L/14~\citep{clip-pmlr-v139-radford21a}, MetaCLIP-L/14~\citep{metaclip_2023demystifying}, and DOVE~\citep{mao2025DOVE}. Discrete baselines include VQGAN~\citep{VQGAN_2021_CVPR}, SD-VQGAN~\citep{SDVQGAN_2022_CVPR}, Llama-Gen~\citep{sun2024llamagen}, TokenFlow~\citep{Tokenflow_2025_CVPR}, VILA-U~\citep{wu2024vila}, TiTok-S~\citep{tiktok_yu2024image}, QLIP~\citep{zhao2025qlip}, and UniTok~\citep{ma2025unitok}.
% To assess image--text understanding, we use Top-1 accuracy. For reconstruction quality, we employ a standard suite of metrics: reconstruction-FID (rFID~\citep{FID_NIPS2017_8a1d6947}), PSNR, SSIM~\citep{SSIM}, and LPIPS~\citep{LPIPS_2018_CVPR}.

% \method demonstrates exceptional reconstruction quality, surpassing all state-of-the-art tokenizers. It achieves an impressive rFID of \textbf{0.31} on ImageNet with a 16$\times$ downsampling ratio. Moreover, \method surpasses all other tokenizers in PSNR, achieving  on the full Datacomp-1B (including 480M resampled images) and  on the 800M subset. These results validate the effectiveness of DQP and DPA's combination. Besides, when expanding training data from 800M to resampled 1B, further reconstruction gains are observed, demonstrating the scalability of \method.
% The comparisons of computation cost are listed in Appendix~\ref{app: compute}.

% Regarding image--text aligned understanding, \method achieves suboptimal, yet competitive, results in zero-shot classification, reaching 70.5\% Top-1 accuracy on ImageNet and notably surpassing UniTok. Initialization significantly impacts performance on the full DC-1B dataset. For example, UniTok improves from 70.8\% to 78.6\% with CLIP pre-trained weights. However, this initialization benefit is marginal for both \method and UniTok on the Resampled DC-1B.

\myparagraph{Reconstruction.}
With CLIP initialization on DC-1B-R, \method achieves a state-of-the-art ImageNet rFID of \textbf{0.31} at $16\times$ spatial downsampling.
Under the same training conditions, \method also improves PSNR from 24.40 to 24.95 dB over the matched UniTok reproduction.
Figure~\ref{fig: reconstruction} shows that \method preserves fine details more faithfully, particularly the lettering on the map and blackboard. Appendix~2 reports reconstruction costs.

% \myparagraph{image--text Alignment.} For zero-shot classification, \method is highly competitive. It achieves 70.5\% Top-1 accuracy on ImageNet, outperforming UniTok. Initialization is crucial for the full DC-1B dataset. For example, UniTok improves from 70.8\% to 78.6\% with CLIP weights. However, this initialization benefit is marginal for both \method and UniTok on the Resampled DC-1B.

\myparagraph{image--text Alignment.}
Moreover, with CLIP initialization on DC-1B-R, \method achieves 70.5\% zero-shot Top-1 accuracy on ImageNet, compared with 69.9\% for the matched UniTok reproduction. In the reported UniTok results on DC-1B, CLIP initialization raises accuracy from 70.8\% to 78.6\%. On the resampled DC-1B-R, the gains from CLIP initialization are smaller for both tokenizers: 1.3 percentage points for UniTok and 0.7 for \method.

% \myparagraph{Data Scaling and Encoder-Only Reference.} Expanding training data from 800M to 1.28B yields only modest gains (66.8\% to 70.1\%) for the randomly initialized model. The extra 480M resampled images have limited diversity, which constrains the performance. Specifically, the model approaches an ``Encoder-Only'' ceiling. This ceiling represents the maximum classification accuracy achievable by the contrastive encoder alone (without quantization). Since our tokenizer's understanding branch learns from this encoder, its accuracy is naturally bounded by the encoder's own capability.

\myparagraph{Data Scaling and Encoder-Only Ceiling.} Increasing training sample presentations from 800M to 1.28B through resampling raises zero-shot Top-1 accuracy from 66.5\% to 69.8\% for randomly initialized \method, and from 66.8\% to 70.1\% for Encoder-Only. Both improve by 3.3 percentage points, with \method remaining 0.3 points below the encoder-only reference in both settings. As Encoder-Only avoids quantization-induced information loss, we use its results in Table~\ref{tab: imagenet} as an empirical ``Encoder-Only'' ceiling for zero-shot classification under random initialization. Since resampling adds no new image--text pairs, the modest gain even for Encoder-Only suggests that the lack of additional data diversity may constrain further improvements.

\begin{table}[t]
\centering
\small
% \fontsize{9}{10.5}\selectfont
% \setlength{\tabcolsep}{0.5mm}
\caption{Image reconstruction on MSCOCO 2017 and ImageNet-1K. CB denotes codebook size. $^\dagger$ indicates CLIP-pretrained initialization; $^\ddagger$ indicates an ImageNet-1K-pretrained tokenizer.}
\label{tab:reconstruction}
\small
\setlength{\tabcolsep}{3pt}
\renewcommand{\arraystretch}{1.15}
\begin{tabular*}{\linewidth}{@{\extracolsep{\fill}}lcrrr|rrr@{}}
\toprule
\multirow{2}{*}{\textbf{Method}}
& \multirow{2}{*}{\textbf{CB}}
& \multicolumn{3}{c|}{\textbf{MSCOCO}}
& \multicolumn{3}{c}{\textbf{ImageNet-1K}} \\
\cmidrule(lr){3-5}
\cmidrule(lr){6-8}
& & \textbf{rFID}
& \textbf{PSNR}
& \textbf{SSIM}
& \textbf{rFID}
& \textbf{PSNR}
& \textbf{SSIM} \\
\midrule
LlamaGen$^{\ddagger}$   & 16K  & 8.40  & 20.28 & 0.55 & 2.47 & 20.65 & 0.54 \\
Show-o                   & 8K   & 9.26  & 20.90 & 0.59 & 3.50 & 21.34 & 0.59 \\
Cosmos                   & 64K  & 11.97 & 19.22 & 0.48 & 4.57 & 19.93 & 0.49 \\
Open-MAGVIT2             & 262K & 6.76  & 22.31 & 0.65 & 1.67 & 22.70 & 0.64 \\
IBQ                      & 262K & 6.79  & 22.28 & 0.65 & 1.53 & 22.69 & 0.64 \\
TokenFlow$^{\dagger}$    & 32K  & --    & --    & --   & 1.37 & 21.41 & 0.69 \\
% \midrule
UniTok                   & 32K  & 4.22  & 23.38 & 0.68 & 0.41 & 23.80 & 0.70 \\
% \textcolor{blue}{UniFlow$^{\dagger}$} & \textcolor{blue}{--} & \textcolor{blue}{\textbf{3.69}} & \textcolor{blue}{\textbf{29.61}} & \textcolor{blue}{\textbf{0.92}} & \textcolor{blue}{0.67} & \textcolor{blue}{\textbf{28.66}} & \textcolor{blue}{\textbf{0.91}} \\
\rowcolor{primitokrow} \primitokrowbackground \textbf{\method}          & 32K  & 3.86 & 24.07 & 0.71 & 0.36 & 24.51 & 0.73 \\
\bottomrule
\end{tabular*}
% \vspace{-0.3cm}
\end{table}

\begin{table*}[t]
\small 
\centering
% \vspace{-0.2cm}
\caption{
\textbf{Visual generation on ImageNet-1K without classifier-free guidance.} $^\dag$VQGAN tokenizer from LlamaGen~\citep{sun2024llamagen}.
$^\ddag$Generated at $384\times384$ and resized to $256\times256$.
$^\star$Visual encoder initialized with pretrained InternViT-300M weights; generated at $448\times448$ and resized to $256\times256$~\citep{yue2025uniflow}.}
\label{tab: generation}
\small
\setlength{\tabcolsep}{3pt}
\renewcommand{\arraystretch}{1.15}
\begin{tabular*}{\linewidth}{@{\extracolsep{\fill}}l c | l c c | cc @{}}
\toprule
 \textbf{Tokenizer} &\textbf{rFID}$\downarrow$ &
\textbf{Generator}   & \textbf{Type} & \textbf{\#Params.} &
% \cmidrule(lr){5-8}\cmidrule(lr){9-12}
\textbf{gFID$\downarrow$} &
\textbf{IS$\uparrow$}
 \\
\midrule
% \multicolumn{8}{c}{\textit{Diffusion Models}}\\
\multirow{2}{*}{SD-VAE~\citep{SDVQGAN_2022_CVPR}} & \multirow{2}{*}{0.61} &
DiT~\citep{DiT_Peebles_2023_ICCV} & Diffusion-based  & 675M & 9.62 & 121.5 \\

& &  SiT~\citep{SiT} & Diffusion-based & 675M & 8.61 & 131.7  \\

% \midrule
% \multicolumn{8}{c}{\textit{Masked Generative Models}}\\
% -  & - & BigGAN~\citep{} & Generative Adversarial & 307M &
% 8.43 & 177.9 \\
Taming-VQGAN$^\dag$~\cite{VQGAN_2021_CVPR}  & 2.19 & LlamaGen$^\ddag$~\citep{sun2024llamagen} & Transformer-based & 1.4B &
  14.65 &    86.3  \\
RQ-VAE~\citep{RQVAE_Lee_2022_CVPR} & 3.20 & RQ-Transformer~\citep{RQVAE_Lee_2022_CVPR}  & Transformer-based & 3.8B & 7.55 & 134.0  \\
ViT-VQGAN~\citep{yu2021vector} & 1.28 & VIM-Large~\citep{yu2021vector}  & Transformer-based & 1.7B & 4.17 & --  \\
 % LFQ~\citep{LFQ_yu2023language}    & 0.90 & MAGVIT-v2~\citep{LFQ_yu2023language} & Transformer-based & 307M &
 % 3.07 & 213.1 \\
% TiTok-L~\citep{tiktok_yu2024image}  & 2.21 & MaskGIT~\citep{MaskGIT_Chang_2022_CVPR} & Transformer-based & 177M &
% 3.15 & 173.0 \\
% \midrule
% \multicolumn{8}{c}{\textit{Autoregressive Models}}\\
MAGE~\citep{MAGE_Li_2023_CVPR} & --& MAGE-ViT~\citep{MAGE_Li_2023_CVPR} & Transformer-based & 230M & 6.93 & 195.8\\
% LFQ~\citep{LFQ_yu2023language}    & 0.90 & MAGVIT-v2~\citep{LFQ_yu2023language} & Transformer-based & 307M &
%  3.07 & 213.1 \\
UniTok~\citep{ma2025unitok} &  0.41 & LlamaGen~\citep{sun2024llamagen} & Transformer-based &  1.4B & 2.51 & 216.7  \\
UniFlow$^\star$~\citep{yue2025uniflow} & -- & MAR-L~\citep{MAR_NEURIPS2024_66e22646} & Transformer-based & 479M & 2.45 & 228.0 \\
\rowcolor{primitokrow} \primitokrowbackground \textbf{\method}   & 0.36 & MAR-L~\citep{MAR_NEURIPS2024_66e22646} & Transformer-based & 427M & 2.85 & 213.6 \\
\bottomrule
\end{tabular*}

% \vspace{-0.2cm}
\end{table*}

\begin{table*}[t]
% \vspace{-0.2cm}
\centering
% \caption{\textbf{Results on Zero-Shot Image-Text Retrieval}. $^\dag$ indicates pretrained CLIP weights initialization. $^*$ denotes our reproduction. Results show that randomly initialized \method outperforms UniTok and even comparable to CLIP initialized QLIP.}

\caption{\textbf{Zero-Shot Retrieval} on Flickr30K and MSCOCO 2014. $^\dag$ indicates initialization with pretrained CLIP weights. $^*$ denotes our reproduction. Results show that randomly initialized \method outperforms UniTok and is even comparable to CLIP-initialized QLIP.}
\label{tab: retrieval}

\footnotesize
\setlength{\tabcolsep}{3pt}
\renewcommand{\arraystretch}{1.15}
\begin{tabular*}{\linewidth}{@{\extracolsep{\fill}}lc c ccc ccc ccc ccc@{}}
\toprule
\multirow{4}{*}{\textbf{Method}}  &\multirow{4}{*}{\textbf{Res.}} & \multirow{4}{*}{\textbf{Data}} & \multicolumn{6}{c}{\textbf{Image to Text}} & \multicolumn{6}{c}{\textbf{Text To Image}} \\
 \cmidrule(lr){4-9} \cmidrule(lr){10-15}
 & && \multicolumn{3}{c}{\textbf{Flickr30K}} & \multicolumn{3}{c}{\textbf{MSCOCO}} & \multicolumn{3}{c}{\textbf{Flickr30K}} & \multicolumn{3}{c}{\textbf{MSCOCO}} \\
 \cmidrule(lr){4-6} \cmidrule(lr){7-9} \cmidrule(lr){10-12} \cmidrule(lr){13-15}
 & & &  \textbf{R@1} & \textbf{R@5} & \textbf{R@10} & \textbf{R@1} & \textbf{R@5} & \textbf{R@10} & \textbf{R@1} & \textbf{R@5} & \textbf{R@10} & \textbf{R@1} & \textbf{R@5} & \textbf{R@10} \\
 % \midrule
   % \multicolumn{15}{c}{\textit{Continuous Tokenizer (CT)}} \\
    \midrule
 % EVA02-B/16 & 224 & 85.7 & 96.7 & 98.9 & 58.7 & 80.7 & 88.2 & 71.2 & 91.0 & 94.7 & 42.2 & 66.9 & 76.3 \\
% CLIP-L/14 &224 & 88.0 & 98.7 & 99.4 & 58.4 & 81.5 & 88.1 & 68.7 & 90.6 & 95.2 & 37.8 & 62.4 & 72.2  \\
EVA02-B/16 & 224 & Merged-2B & 87.9 & 97.8 & 99.2 & 60.0 & 82.7 & 89.1 & 73.9 & 91.1 & 94.7 & 42.2 & 67.5 & 77.1 \\
CLIP-L/14 &224 & WIT400M & 86.4 & 97.3 & 99.4 & 57.1 & 80.1 & 87.2 &  67.9 & 88.9 & 93.3 & 35.4 & 60.4 & 71.0  \\
 \midrule
  % \multicolumn{15}{c}{\textit{Discrete Tokenizer (DT)}} \\
   % \midrule
%  UniTok$^{*}$ & 256 & DC-800M  & 73.8 &   93.1  &   95.5   &    47.9  &  73.0    &  80.7    &   56.1   &   81.2   &   87.6   & 30.9 &   55.2  &  66.3    \\
% \rowcolor{lightblue}
% \textbf{\method} &  256 & DC-800M  &  73.8  & 92.1    & 95.6    &   48.9  &   73.6   &   81.9   &   55.1   &   80.1   &   87.3   &  31.9    &   56.2   &  66.8    \\
% $\text{VILA-U}^{\dag}$  &256&  COYO700M &    &      &      &     &      &      &      &      &      &      &      &      \\
 % \midrule
 $\text{QLIP}^{\dag}$  & 256&   DC-1B & 80.4  &  94.0 &   97.0   &   50.6   &    74.1  &  82.2    &   61.8   &   84.5   &   89.9   &  32.9    &   57.2   &   67.6   \\
  UniTok  & 256 & DC-1B &  79.5   &   95.6   &   97.3   &  52.4    &   76.4   &  84.4    &   60.8   &  84.4    &   90.3   &   34.8    &   59.6   &  69.9   \\
UniTok$^{*}$ & 256& DC-1B-R  & 75.8 &  93.3    &   96.6   &   50.9   &     74.5 &  83.4    &  57.9    &   82.8   &   89.0   &  33.2    &  57.6    &  68.2    \\
\rowcolor{primitokrow} \primitokrowbackground \textbf{\method} &  256 & DC-1B-R & 78.7 &   93.3   &   96.5   &  53.0    &     75.4 & 83.8     &   59.3   &  82.8   &  89.0    &    34.0  &  58.8    &  69.1    \\
% UniTok$^{\dag}$ & 256& DC-1B-R  &  &      &   &      &     &    &    &    &    &    &    &     \\
% \rowcolor{lightblue}
% \textbf{\method}$^{\dag}$ &  256 & DC-1B-R &  &      &      &      &      &      &      &      &      &      &      &      \\
\bottomrule
  \end{tabular*}

\end{table*}

\subsection{Quantitative Evaluation}

\myparagraph{Cross-dataset reconstruction.}
As shown in Table~\ref{tab:reconstruction}, \method outperforms the original DC-1B UniTok model on both reconstruction benchmarks.
On MSCOCO 2017, \method reduces rFID from 4.22 to 3.86 and improves PSNR by 0.69 dB. We observe a similar gain on ImageNet-1K, where rFID falls from 0.41 to 0.36 and PSNR increases by 0.71 dB.

% \myparagraph{Multimodal Understanding.} We evaluate \method in LLaVA-1.5~\citep{liu2024improved} with Vicuna-v1.5-7B. \method achieves the highest reported average score of 65.2 among the compared tokenizers with complete results in Table~\ref{tab:multimodal_understanding}. Compared with the DFN-CLIP-L variant of UniFlow-LV, it improves MM-Vet from 30.2 to 32.8 and MME-S from 1748.0 to 1787.4. It also improves GQA and MME-P and matches MMBench, while obtaining lower scores on POPE and TextVQA.

\myparagraph{Multimodal Understanding.}
We evaluate \method in LLaVA-1.5~\citep{liu2024improved} with Vicuna-v1.5-7B. Table~\ref{tab:multimodal_understanding} reports an average score of 65.4 for \method. Our tokenizer is trained from random initialization, whereas the compared UniFlow-LV variant initializes its visual encoder with pretrained DFN-CLIP-L weights. Compared with this variant, \method improves MM-Vet from 30.2 to 32.8 and MME-S from 1748.0 to 1787.4. It also improves GQA and MME-P and matches MMBench.

% Sources: UniTok arXiv:2502.20321v3, Table 3; UniFlow arXiv:2510.10575v2, Table 2 and Appendix B.2.
% TODO: Replace the \method dashes below with measured benchmark results before making comparative performance claims.

\myparagraph{Visual Generation.}
We evaluate visual generation on ImageNet-1K at $256\times256$ resolution without classifier-free guidance. We train MAR-L~\citep{MAR_NEURIPS2024_66e22646} from scratch on ImageNet-1K using \method as the image tokenizer. Table~\ref{tab: generation} shows that \method achieves a gFID of 3.06 and an IS of 213.6, compared with 2.51/216.7 for UniTok--LlamaGen~\citep{ma2025unitok} and 2.45/228.0 for UniFlow--MAR-L~\citep{yue2025uniflow}.

While our results remain below these baselines, UniFlow trains a larger MAR-L model with a batch size of 1024, whereas we use a global batch size of 256 due to computational constraints. UniFlow also generates images at $448\times448$ before resizing them to $256\times256$ and initializes its tokenizer encoder with pretrained InternViT-300M weights. Our tokenizer is trained from random initialization. These differences limit attribution of the performance gap to tokenizer design alone.

 % \noindent \textcolor{red}{\myparagraph{image--text Retrieval.} On the 800M Datacomp-1B subset, \method surpasses UniTok in accurately retrieving image--text pairs, showing strong cross-modal alignment. However, due to training constraints and information loss from vector quantization, it is not yet competitive with non-quantized CLIP models. Details is listed in Appendix~\ref{app: retrieval}.}

\myparagraph{Zero-Shot image--text Retrieval.}
Under DC-1B-R training, \method improves R@1 over the matched UniTok reproduction in both retrieval directions on Flickr30K and MSCOCO, as shown in Table~\ref{tab: retrieval}. Image-to-text R@1 increases from 75.8 to 78.7 on Flickr30K and from 50.9 to 53.0 on MSCOCO, while text-to-image R@1 rises from 57.9 to 59.3 and from 33.2 to 34.0, respectively. Overall, these results support stronger cross-modal alignment of \method under matched training conditions.

\begin{figure*}[t]
\centering
% \vspace{-0.5cm}
\includegraphics[width=0.8\linewidth]{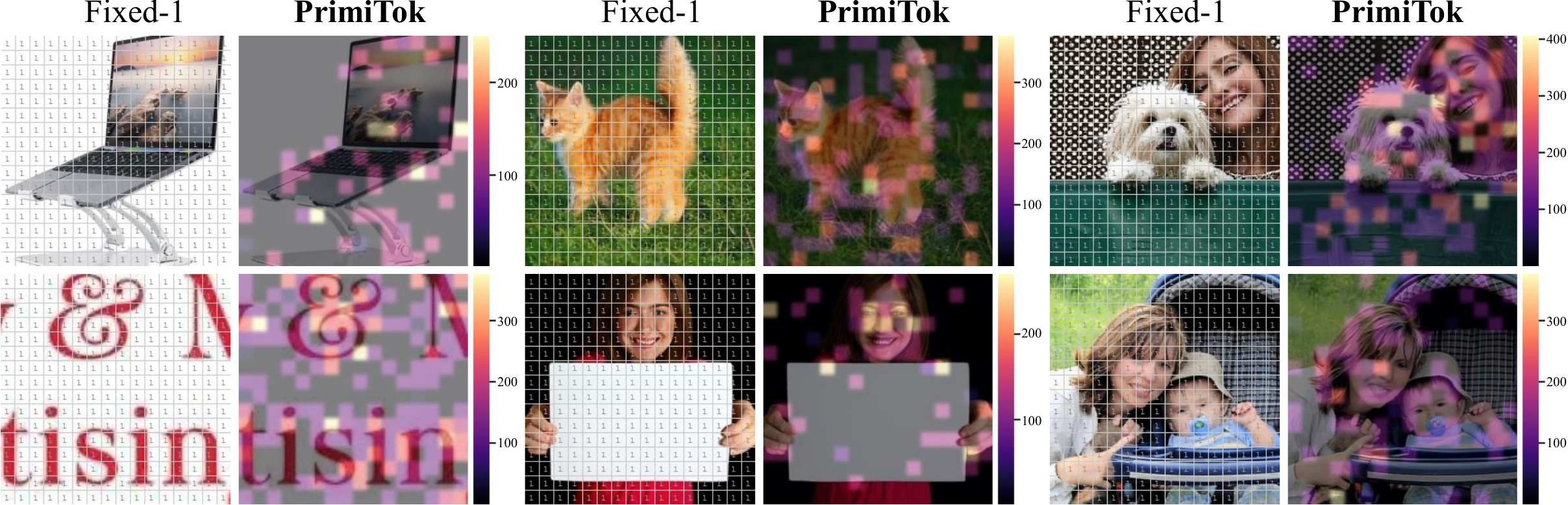}
% \vspace{-0.2cm}
% \caption{\textbf{DPA Effectiveness.} Fixed-1 assigns one primitive per token chunk, yielding uniform allocation. \method equipped with DPA produces an adaptive heatmap, focusing more primitives on areas with complex information. Color bars denote primitive count.}
\caption{
\textbf{Content-Adaptive Primitive Allocation.}
Fixed-1 assigns one primitive per token chunk, whereas \method tends to allocate more primitives to regions containing fine details and salient image content. Brighter colors indicate larger primitive counts.
}
\label{fig: ablation_filtering}

\end{figure*}

\begin{table}[t]
\small 
% \vspace{-0.3cm}
\centering
\caption{\textbf{Ablation Studies on the ImageNet-1K Validation Set.} Bold denotes the default configuration.}
\label{tab:comprehensive_ablation}
\small
\setlength{\tabcolsep}{3pt}
\renewcommand{\arraystretch}{1.15}
\begin{tabular*}{\linewidth}{@{\extracolsep{\fill}}l c c c c c@{}}
    \toprule
    \textbf{Setting} & 
    \textbf{Accuracy} & 
    \textbf{rFID$\downarrow$} & 
    \textbf{PSNR$\uparrow$} & 
    \textbf{SSIM$\uparrow$} & 
    \textbf{LPIPS$\downarrow$} \\
       \midrule
    \multicolumn{6}{c}{\textit{Warm-Up Training Strategy}} \\
    \midrule
    w/o Warm-up        & 63.8 & 0.34 & 24.45 & 0.72 & 0.1206 \\
    \textbf{w/ Warm-up}     &   \textbf{66.5} & \textbf{0.37} & \textbf{23.93} & \textbf{0.71} & \textbf{0.1410} \\
    \midrule
    \multicolumn{6}{c}{\textit{Primitive Allocation}} \\
    
    \midrule
    Fixed-1   & 65.8 & 0.41 & 23.57 & 0.69 & 0.1440 \\
    Fixed-60     & 63.9 & 0.39 & 23.61 & 0.68 & 0.1420  \\
    \textbf{with DPA} & \textbf{66.5} & \textbf{0.37} & \textbf{23.93} & \textbf{0.71} & \textbf{0.1410} \\
    \midrule
    \multicolumn{6}{c}{\textit{Top-$K$ Candidate Selection for Allocator $\mathcal{A}$}} \\
    \midrule
    500            & 65.5 & 0.94 & 23.74 & 0.70 & 0.1502 \\
    \textbf{1000}     &     \textbf{66.5} & \textbf{0.37} & \textbf{23.93} & \textbf{0.71} & \textbf{0.1410} \\
    Vocabulary Size        & 64.7 & 0.34 & 24.23 & 0.71 & 0.1385 \\
    \midrule
    \multicolumn{6}{c}{\textit{Sub-Codebook Diversity Regularization ($\mathcal{L}_{\text{DQP}}$)}} \\
    \midrule
    w/o Diversity      & 66.1 & 0.43 & 23.82 & 0.70 & 0.1451 \\
    \textbf{w/ Diversity}    &  \textbf{66.5} & \textbf{0.37} & \textbf{23.93} & \textbf{0.71} & \textbf{0.1410} \\
    \bottomrule
\end{tabular*}
% \vspace{-0.5cm}
\end{table}

\subsection{Ablation Studies and Analysis}
\label{sec: ablation}

Table~\ref{tab:comprehensive_ablation} examines warm-up training, primitive allocation, candidate-pool size, and sub-codebook diversity regularization. Unless varied in the corresponding ablation, the default configuration uses a 25\% warm-up, DPA with $K=1000$, and diversity regularization.

\begin{figure}[t]
    \centering
    % \vspace{-0.3cm}
    \begin{subfigure}[t]{0.4\linewidth}
        \centering
        \includegraphics[width=\linewidth]{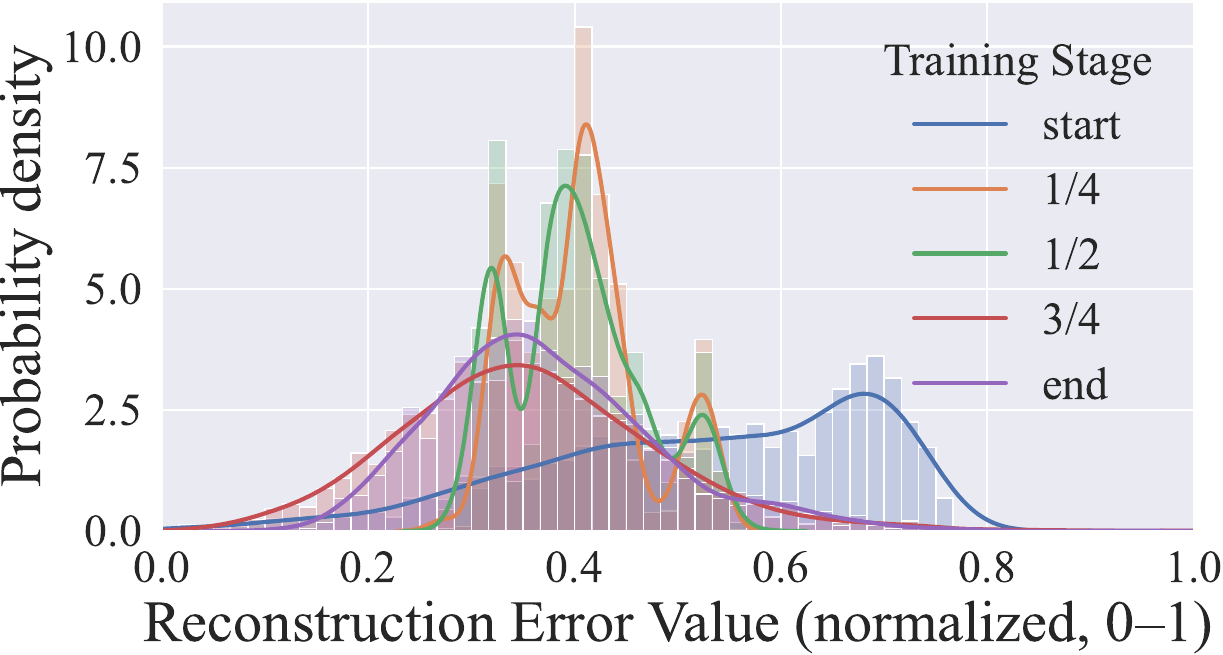}
        \caption{Reconstruction Error during Training.}
        \label{fig:ablation_training}
    \end{subfigure}
    \hfill
    \begin{subfigure}[t]{0.55\linewidth}
        \centering
        \includegraphics[width=\linewidth]{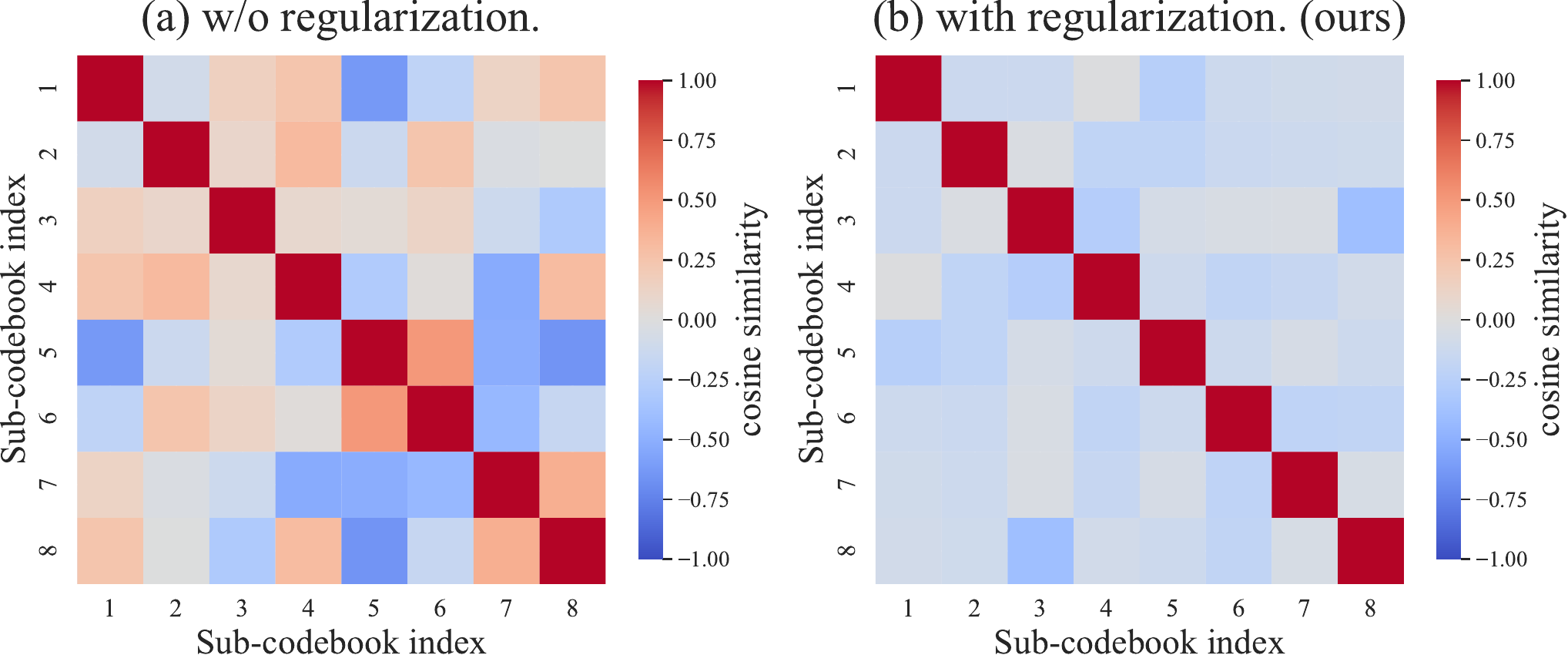}
        \caption{Pairwise Similarity of Sub-Codebook Centroids.}
        \label{fig:ablation_diversity}
    \end{subfigure}
    \vspace{-0.3cm}
\end{figure}

\myparagraph{Warm-Up Training.}
Since DPA relies on reconstruction errors to estimate token complexity, its allocation signals may be unreliable during early training, when latent representations are still evolving. Figure~\ref{fig:ablation_training} illustrates the evolution of reconstruction-error distributions across training stages. Consistent with this observation, Table~\ref{tab:comprehensive_ablation} shows that a 25\% warm-up improves Top-1 accuracy from 63.8\% to 66.5\%, at a slight cost to reconstruction quality. We therefore adopt this warm-up period, delaying DQP regularization and DPA-based allocation to favor semantic alignment while maintaining reconstruction performance.

\myparagraph{Content-Adaptive Primitive Allocation.}
Table~\ref{tab:comprehensive_ablation} compares DPA with fixed allocation under the same training setting. Measured on the ImageNet validation set, DPA uses an average of approximately 60 primitives per token chunk. We therefore include Fixed-60 as a uniform-allocation baseline with a comparable average primitive count. DPA outperforms both fixed-allocation baselines across all five metrics. Compared with Fixed-60, it improves Top-1 accuracy from 63.9\% to 66.5\% and reduces rFID from 0.39 to 0.37, supporting the benefit of content-adaptive allocation beyond simply increasing the primitive count. 

Figure~\ref{fig: ablation_filtering} compares the primitive allocations, showing uniform allocation in Fixed-1 and spatially varying allocation in \method. Since latent tokens incorporate spatial context, these heatmaps represent token-level allocation rather than direct measures of local image complexity.

\myparagraph{Top-$K$ Candidate Selection for Allocator $\mathcal{A}$.}
Among the candidate pools in Table~\ref{tab:comprehensive_ablation}, $K=1000$ yields the highest Top-1 accuracy of 66.5\%, with an rFID of 0.37. Reducing the pool to $K=500$ lowers accuracy to 65.5\% and raises rFID to 0.94. Using the full vocabulary further improves rFID to 0.34, but accuracy falls to 64.7\%. We use $K=1000$ for its classification advantage and competitive reconstruction quality.

% \begin{figure}[t]
%   \centering
%   \vspace{-0.3cm}
%     \includegraphics[width=0.8\linewidth]{fig/loss_crop.pdf}
%         \vspace{-0.1cm}
%     \caption{Reconstruction Error during Training.}
%     \label{fig:ablation_training}
%  \vspace{-0.3cm}
% \end{figure}

%   \begin{figure}[t]
%     \centering
%      \vspace{-0.3cm}
%     \includegraphics[width=1.0\linewidth]{fig/diversity_crop.pdf}
%         \vspace{-0.2cm}
%     \caption{Pairwise Similarity of Sub-Codebook Centroids.}
%     \label{fig:ablation_diversity}
%     \vspace{-0.5cm}
%   \end{figure}

% \myparagraph{Sub-Codebook Diversity.}
% Table~\ref{tab:comprehensive_ablation} shows that adding the sub-codebook diversity regularization loss $\mathcal{L}_{\mathrm{DQP}}$ improves Top-1 accuracy from 66.1\% to 66.5\% and reduces rFID from 0.43 to 0.37, with further improvements in PSNR, SSIM, and LPIPS. Figure~\ref{fig:ablation_diversity} shows lower pairwise similarity between sub-codebook centroids with regularization. Together, these results support the contribution of diversity regularization to both reconstruction and recognition.

\myparagraph{Sub-Codebook Diversity.}
Table~\ref{tab:comprehensive_ablation} demonstrates the effectiveness of DQP in reducing inter-codebook redundancy. Adding the centroid-level diversity regularization $\mathcal{L}_{\mathrm{DQP}}$ improves Top-1 accuracy from 66.1\% to 66.5\% and reduces rFID from 0.43 to 0.37, with consistent gains in PSNR, SSIM, and LPIPS. Figure~\ref{fig:ablation_diversity} further shows reduced pairwise cosine similarity among sub-codebook centroids, indicating more diverse primitive representations.

\section{Conclusion}
This work introduces \method, a compositional visual tokenizer designed to unify understanding and generation. It follows an encoder–quantizer–decoder framework, with DQP and DPA as its core quantization components. DQP encourages sub-codebook diversity through centroid-level regularization, helping limit redundancy among primitive groups. DPA adapts primitive allocation to mitigate information loss during tokenization. Experiments demonstrate the effectiveness of adaptive primitive composition in preserving visual fidelity and semantic information.

% ── References ───────────────────────────────────
\bibliographystyle{unsrtnat}
\bibliography{reference}

\end{document}